\documentclass[runningheads]{llncs}
\usepackage[T1]{fontenc}
\usepackage{graphicx}
\usepackage{latexsym}
\usepackage{amssymb}
\usepackage{amsmath}
\usepackage{booktabs}
\usepackage{enumitem}
\usepackage{graphicx}
\usepackage{color}
\usepackage{tabularx} 
\usepackage{booktabs} 
\usepackage{calc}     
\usepackage[table]{xcolor}
\usepackage{multirow} 
\usepackage{boldline}
\usepackage{hyperref} 
\usepackage{subcaption} 
\usepackage{float}
\begin{document}
\title{TSI: A Multi-View Representation Learning Approach for Time Series Forecasting}
%
%

\author{Wentao Gao\inst{1}\orcidID{0009-0009-8945-2946} \and
Ziqi Xu\inst{2}\orcidID{0000-0003-1748-5801} \and
Jiuyong Li\inst{1}\orcidID{0000-0002-9023-1878} \and
Lin Liu\inst{1}\orcidID{0000-0003-2843-5738} \and
Jixue Liu\inst{1}\orcidID{0000-0002-0794-0404} \and
Thuc Duy Le\inst{1}\orcidID{0000-0002-9732-4313} \and
Debo Cheng\inst{1}\orcidID{0000-0002-0383-1462} \and
Yanchang Zhao\inst{3}\orcidID{0000-0002-0209-3971} \and
Yun Chen\inst{3}\orcidID{0000-0001-6819-6562}}
\authorrunning{Wentao Gao, Ziqi Xu et al.}
%
\institute{University of South Australia, SA 5095, AU
\email{gaowy014@mymail.unisa.edu.au}\and
CSIRO, Melbourne, VIC, AU
\email{ziqi.xu@data61.csiro.au}\and
CSIRO, Canberra, ACT, AU
\email{\{yanchang.zhao, yun.chen\}@data61.csiro.au}}

%
%
\maketitle              
\begin{abstract}
As the growing demand for long sequence time-series forecasting in real-world applications, such as electricity consumption planning, the significance of time series forecasting becomes increasingly crucial across various domains. This is highlighted by recent advancements in representation learning within the field. This study introduces a novel multi-view approach for time series forecasting that innovatively integrates trend and seasonal representations with an Independent Component Analysis (ICA)-based representation. Recognizing the limitations of existing methods in representing complex and high-dimensional time series data, this research addresses the challenge by combining TS (trend and seasonality) and ICA (independent components) perspectives. This approach offers a holistic understanding of time series data, going beyond traditional models that often miss nuanced, nonlinear relationships. The efficacy of TSI model is demonstrated through comprehensive testing on various benchmark datasets, where it shows superior performance over current state-of-the-art models, particularly in multivariate forecasting. This method not only enhances the accuracy of forecasting but also contributes significantly to the field by providing a more in-depth understanding of time series data. The research which uses ICA for a view lays the groundwork for further exploration and methodological advancements in time series forecasting, opening new avenues for research and practical applications.
\keywords{Time series forecasting \and Representation learning \and ICA.}
\end{abstract}

\section{Introduction}

Time series forecasting holds a pivotal role in machine learning and statistical analysis, particularly salient in domains such as financial market analytics \cite{FinancialMarkets2022}, meteorological prognostication \cite{Meteorology2021}, and the forecasting of energy demands \cite{EnergyForecasting2022}. For instance, within the ambit of meteorological forecasting, the precision of time series predictions is paramount in mitigating the impacts of natural catastrophes, including torrential rains, droughts, and tempests \cite{NaturalDisasters2021}. These phenomena, representing significant global challenges, profoundly affect agriculture, water resource management, and ecological systems \cite{ClimateImpact2020}. Concomitant with the exacerbation of global climate changes, the development of efficacious methodologies for the prediction of these varied natural calamities through time series analysis has emerged as an exigent imperative \cite{ClimateChange2023}.

Traditional time series prediction methods like ARIMA \cite{boxjenkins1960arima}, ETS model \cite{brown1956exponential} and Wavelet Transform model \cite{daubechies1992wavelets}, often struggle to handle the nonlinear features and dynamic changes of time series data. Emerging machine learning based prediction models provide new approaches to tackle these challenges, including transformer based models like Informer \cite{haoyietal-informer-2021}, FEDformer \cite{zhou2022fedformer}. Yet, transformer models, despite their proficiency in many areas like weather\cite{gao2024deconfoundingapproachclimatemodel}, often fall short in explicitly delineating the underlying dynamics they capture, an area where representation learning methods show greater potential.

An alternative approach to improve prediction performance in time series forecasting involves learning representations. Commonly, a representation learning based method decomposes time series data into trend and seasonal components \cite{yue2022ts2vec}\cite{woo2022cost}, effectively capturing patterns and predictive information. However, these methods might miss crucial information in complex, high-dimensional data due to their inherent representational limitations. To counter this, Independent Component Analysis (ICA) is used to extract independent source representations, revealing subtle and nonlinear relationships in the data, thus capturing key hidden features \cite{financialICA}\cite{icaSvrForecasting}. This ICA method addresses the gaps in traditional time series analysis, offering deeper insights for forecasting.

\begin{table}[h]
\caption{Trend/Seasonal Decomposition vs ICA}
\centering
\begin{tabular}{|l|p{4.5cm}|p{4.5cm}|}
\toprule
 & \textbf{Trend/Seasonal} & \textbf{ICA} \\
\midrule
\textbf{Focus} & Periodicity & Hidden structures \\
\textbf{Strength} & Intuitive; Ideal for clear cycles & Unveils complex, non-linear patterns \\
\textbf{Limitation} & Overlooks non-cyclic features & Computationally intensive \\
\bottomrule
\end{tabular}
\label{tab:three_line_comparison}
\end{table}

The above described trend and seasonal decomposition approach maps data into latent spaces based on set patterns, which may have the risk of mislabelling key information as noise due to their reliance on repetitive cycles. Conversely, ICA effectively identifies distinct data components but may miss specific patterns, such as trends and seasonality. As Table 1 suggests, these are complementary. Despite this, their integration has not been explored in literature.

This study introduces an innovative hybrid of TS and ICA. By fusing trend, seasonal, and independent elements recognized through ICA, we have developed a multi-view time series forecasting model. This model captures both overarching trends and complex nonlinear dynamics. In comparison with contemporary advanced methods on several benchmark datasets, our approach demonstrates enhanced performance in time series representation and forecasting.

The main contributions of this study include:

\begin{enumerate}
    \item Introducing a novel multi-view approach for time series forecasting which integrates trend and seasonal representation with an ICA derived representation. This approach provides a holistic understanding of time series data, encapsulating a comprehensive perspective on temporal data characteristics and complex non-linear  patterns for enhanced predictive accuracy.
    \item The efficacy and broad applicability of the TSI model are evaluated through experimentation across a diverse array of benchmark datasets. The proposed method exhibits superior performance, surpassing current state-of-the-art models in multivariate time series forecasting.
\end{enumerate}


\section{TSI Representation Learning Approach}

\subsection{Problem formulation}

In this study, we examine a time series of dimension $m$, denoted as $(\mathbf{x}_1, \ldots, \mathbf{x}_T) \in \mathbb{R}^{T \times m}$. Our objective is to use historical data spanning $h$ steps to predict the future $k$ steps of the time series, denoted as ${Y}^{*} = g(X)$, where $X \in \mathbb{R}^{h \times m}$ represents the input historical data and ${Y}^{*} \in \mathbb{R}^{k \times m}$ represents the predicted future values. In this research, we focus on enhancing the performance of the predictive function $g(\cdot)$ by extracting deep feature representations $H = f(X)$ from the historical data $X$, where $H \in \mathbb{R}^{h \times d}$ maps the $m$-dimensional raw signals into a $d$-dimensional latent space at each timestamp.

To achieve this goal, we develop a nonlinear embedding function $f(\cdot)$ that not only captures the complex patterns within the historical data but also enhances the predictive model $g(\cdot)$ by integrating advanced feature representations $H$, thereby improving predictions for future time steps. Specifically, we utilize the final step feature representation $H_h$ from the historical data as an enriched input to the predictive model, allowing predictor to make more accurate predictions based on a rich contextual understanding. This approach not only optimizes the feature extraction process but also improves the overall predictive framework's performance through refined feature representations.

\begin{figure}[h]
    \centering
    \includegraphics[width=0.5\linewidth]{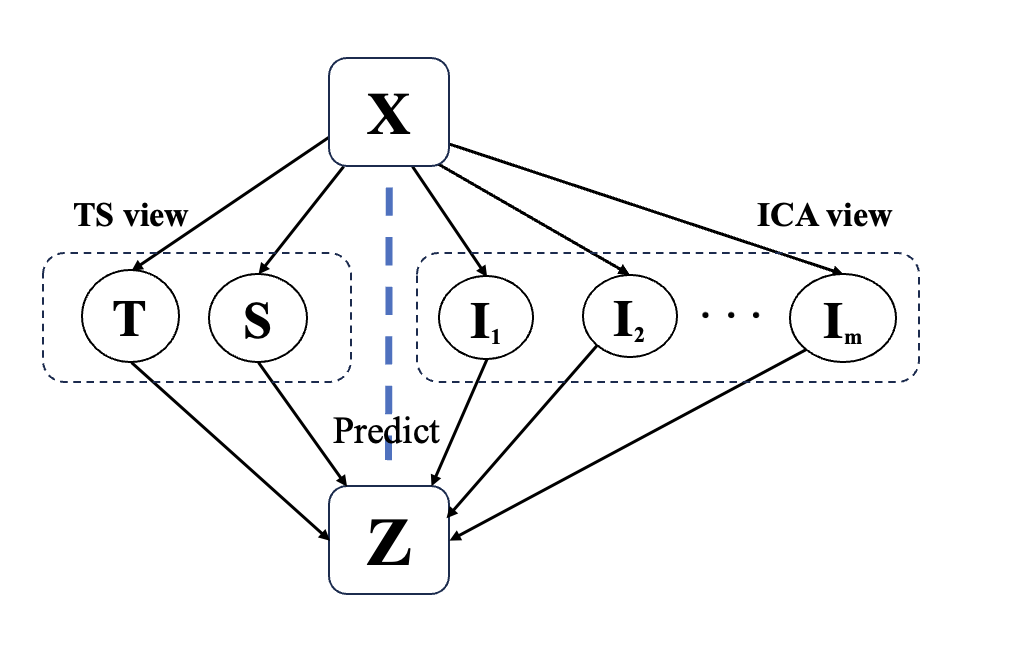}
    \caption{Schematic of TSI, the proposed multi-view approach for time series forecasting}
    \label{fig:enter-label}
\end{figure}

\subsection{TSI feature representation}
 In the realm of time series analysis, unraveling the intricate patterns and underlying factors influencing trends is crucial for accurate forecasting and interpretation. The TSI feature Decomposition approach integrats two powerful analytical perspectives: Trend Seasonality (TS) analysis and Independent Component Analysis (ICA). This multi-view method is shown in Figure 1 aims to provide a more nuanced and detailed understanding of time series data, especially in contexts where complex environmental and climatic factors play a significant role. By integrating the broad, overarching insights offered by TS analysis with the granular, independent factors revealed through ICA, this approach not only captur the 'essence of the data,' which includes fundamental components such as prominent trends and seasonality, and delves deeper into hidden patterns through ICA’s ability to separate high-order statistical dependencies, such as subtle yet impactful cyclical fluctuations and random perturbations.

\begin{figure*}[htbp]
    \centering
    \includegraphics[width=0.8\linewidth]{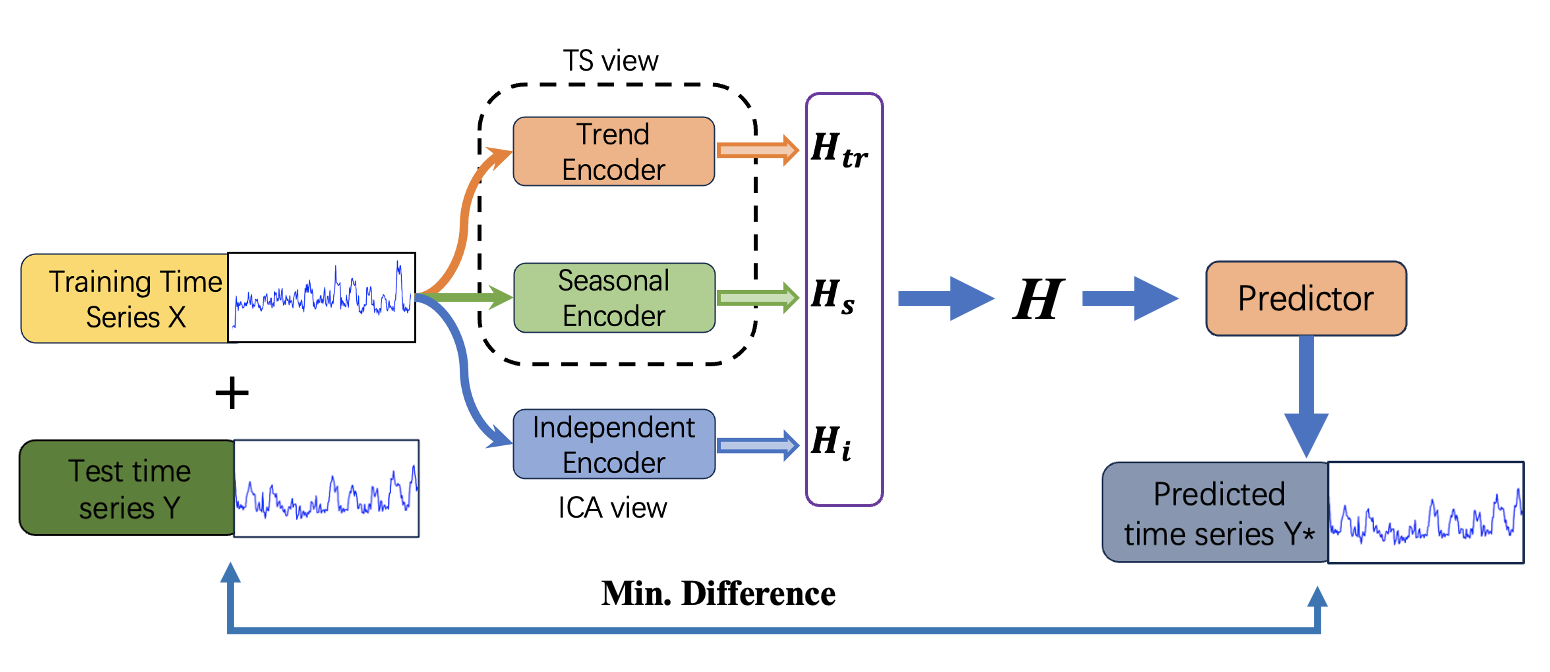}
    \caption{The proposed multi-view time series forecasting model, incorporating Trend, Seasonal, and ICA encoders. The model's objective is to obtain a forecasted time series \( Y^* \) that has the smallest difference from the test time series \( Y \).}
    \label{fig:enter-label}
\end{figure*}

\textbf{ICA view}. Independent components (denoted as \(  I_1, \ldots, I_m \)) directly recover from the observed time series data (\( X \)). These independent factors represent fine-grained, high-frequency variations that are typically not captured by broader trend or seasonal patterns \cite{Forootan2018CICA}. ICA decomposes the time series into statistically independent components \cite{Hyvarinen2000ICA},  uncovering potential latent variables that generate the observed data. For example, these variables are often associated with specific environmental or climatic factors, such as natural phenomena or human activities. 

\textbf{TS view}. By isolating the trend (\( T \)) and seasonality (\( S \)) components, significant long-term changes in the time series can be interpreted as responses to latent factors. For instance, a consistent upward trend might reflect the prolonged impact of global warming, while regular seasonal shifts could be linked to cyclical climate changes. These insights at various views provide a backdrop for identifying relationships and attributing certain changes or patterns to climate shifts or natural cycles.

\textbf{Objective}. By integrating TS and ICA analyses, we establish a multi-view approach. This framework combines the TS view on overarching trends and seasonal patterns with the detailed exploration of independent components provided by the ICA view, enabling a comprehensive understanding and prediction of complex dynamics within the data. Together, these will enhance our understanding of the mechanisms behind these predictions.

\subsection{The proposed multi-view approach}

In this section, we introduce the TSI, which is focused on enhancing the diversity and robustness of feature representation in time series analysis.

The multivariate time series data \( X \), is first decomposed into its constituent components: the Trend \( H_{tr} \), Seasonality \( H_s \), and Independent Components \( H_i \).

Following the schematic of TSI shown in Figure 1, the proposed overall process for feature representation learning with TSI is presented in Figure 2. \(H\) is conceptualized as the aggregate of trend, seasonality, and independent components. The comprehensive feature representation can be formulated as follows:
\begin{equation}
    H = [H_{tr}; H_s; H_i]
\end{equation}
where \( [;] \) denotes the concatenation operation.

After we have obtained a well-trained feature representation H
H, which captures the nonlinear relationships of the original data, we can directly apply this representation to linear regression for predictive purposes. 
To predict future $k$ steps values $Y^*$ within the last time step \(t\)'s representation $H_{t}$, we employ the following model:
\begin{equation}
\mathbf{Y^*}_{t+1:t+k} = f(H_{t})
\end{equation}
where $f(\cdot)$ denotes a linear regression model that incorporates an L2 regularization term, utilizing the feature representation $H_{t}$ as its input to forecast the forthcoming values $\mathbf{Y^*}$. Specifically, this approach leverages the strength of linear regression models in handling continuous data predictions, while the L2 regularization term helps mitigate the risk of overfitting by penalizing large coefficients, ensuring a more generalized model performance. 

\begin{figure*}[!h]
    \centering
    \includegraphics[width=1\linewidth]{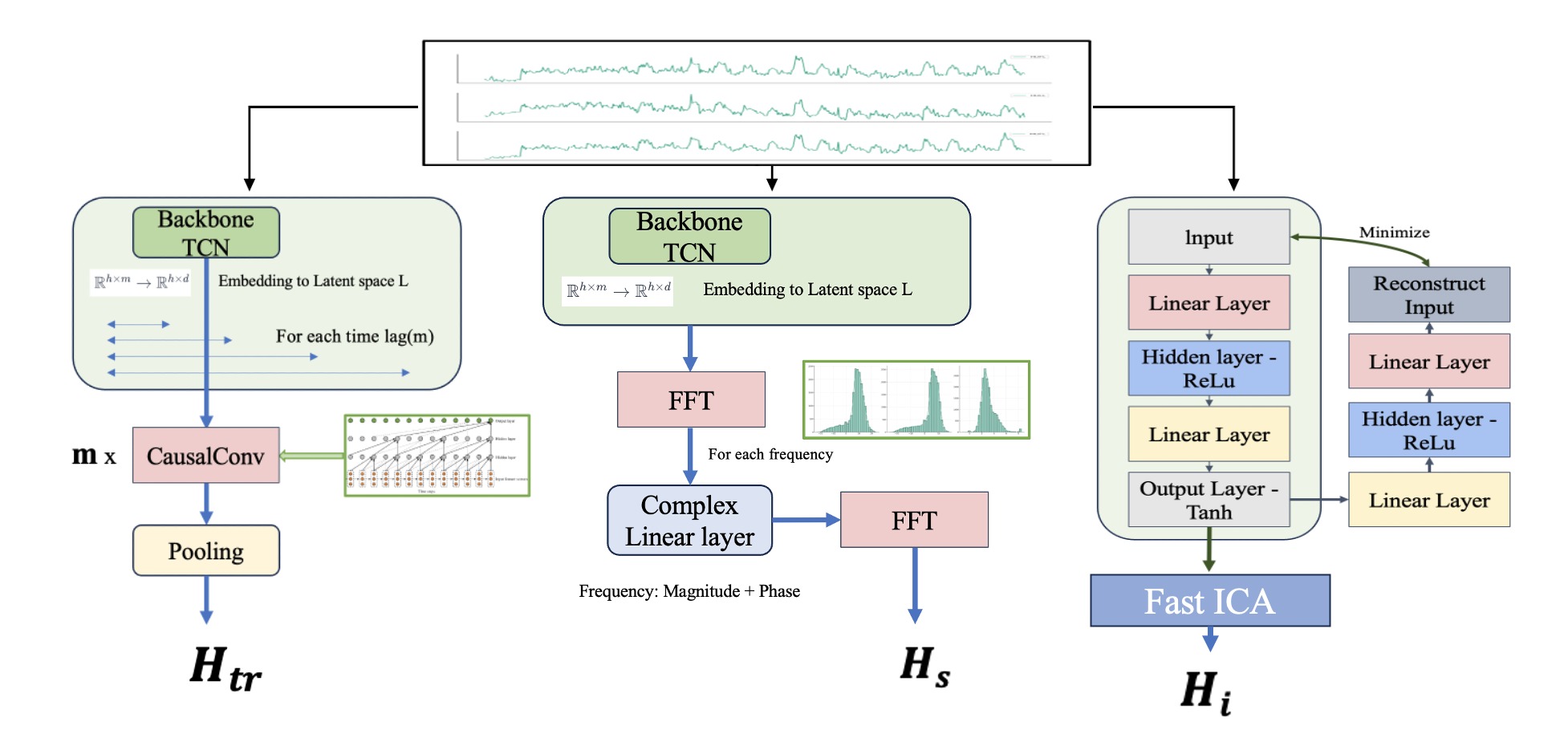}
    \caption{Architectural Overview of the Time Series Decomposition Model. The left block extracts the trend component \( H_{tr} \) using a Temporal Convolutional Network (TCN) and pooling. The middle block captures the seasonal component \( H_{s} \) via FFT and a complex linear layer to encode frequency and phase. The right block extracts the independent component \( H_{i} \) using a fully connected network with activations designed to minimize reconstruction error. This model decomposes time series into distinct features for robust representation.}
    \label{fig:enter-label}
\end{figure*}

\subsection{Trend \& Seasonal Representation}
we are obtain trend and seasonality as most time series representation work does. 
\begin{equation}
H_{tr} = \frac{1}{M+1} \sum_{j=0}^{M} \Omega(G_{j}, 2^j)
\end{equation}

Here, \( \Omega \) denotes the specialized causal convolution function\cite{oord2016wavenet}. The final output \( H_{tr} \) emerges as a synthesis of contributions from individual layers, averaged to formulate a composite yet distinctive representation of the time series trend.

To further refine our model and bolster its ability to distinguish different patterns, we employ a time-domain contrastive loss, inspired by the Momentum Contrast (MoCo) framework\cite{he2020momentum}, which enhances our feature representations by effectively differentiating positive and negative sample pairs. 

For robust seasonal feature extraction in time series analysis, we adopt a Fourier Transform (FT)-based decomposition, as suggested by Oppenheim et al. (2009) and recently by \cite{zhou2022fedformer}\cite{woo2022cost}. This technique facilitates the dissection of time series into constituent seasonal components by projecting the data into the frequency domain.

Building upon \cite{woo2022cost}, a learnable Fourier transformation layer is introduced to encourage nuanced interactions between different frequency components. This is achieved by assigning a distinctive set of complex-valued weights to each frequency, allowing for a tailored enhancement of the seasonal patterns present in the time series data. Our framework integrates this technique into our model's architecture.

The frequency domain interactions and subsequent transformations can be expressed through the following formulation:

\begin{equation}
(H_{s})_{i,l} = \text{IFT}\left\{ \sum_{n=1}^{N} P_{i,n,l} \odot \text{FT}\{Q_{i,n}\} + B_{i,l} \right\}
\end{equation}

Here, \( (H_{s})_{i,l} \) represents the matrix element of the extracted seasonal feature at the \( i \)-th observation and \( l \)-th frequency component. The operators \( \text{IFT} \) and \( \text{FT} \) denote the Inverse Fourier Transform and Fourier Transform respectively. \( P_{i,n,l} \) symbolizes the transformation coefficients tailored to each frequency, while \( B_{i,l} \) corresponds to the bias term incorporated within the transformation. The original time series data before transformation is \( Q_{i,n} \), with \( N \) being the number of elements considered. 

When implementing trend and seasonal encoder, we apply time-domain contrastive loss to each time series sample.In the context of our study, we adopted a triad of data augmentation strategies, namely scaling, shifting, and jittering\cite{yue2022ts2vec}. Each technique is probabilistically activated with a chance of 50\%. More detailed information are described in source code \url{https://github.com/Wentao-Gao/TSI-forcasting}.

\subsection{Independent representation}
From the ICA perspective, we believe that independent representations offer a more detailed and nuanced understanding of data dynamics compared to the broader insights provided by the TS view.

This study explores the application of nonlinear Independent Component Analysis (nICA) for extracting latent, independent sources from high-dimensional datasets. Traditional linear ICA approaches often struggle with datasets characterized by complex, nonlinear interdependencies, underscoring the need for nonlinear mappings for effective source separation.

In our framework, we assume the time series data adhere to the general nonlinear mixing model as defined by \cite{Hyvarinen2000ICA}. Here, \( X_m \) represents the \( m \)-th observed variable among \( n \) total variables, expressed as:

\begin{equation}
     X_m = f_m \left( \sum_{j=1}^{n} a_{mj} I_j \right), \quad m = 1, \ldots, n
\end{equation}

Here, \( m \) indexes the observed variables, with \( I_j \) denoting the source signals and \( a_{mj} \) the mixing coefficients. The source signals \( I_j \), for \( j = 1, 2, \ldots, n \), undergo linear mixing followed by a transformation through the nonlinear function \( f_m \) to produce the observed variables \( X_m \).

To approximate this nonlinear mapping, we utilize deep neural networks within the framework of Variational Autoencoders (VAEs). VAEs are known for their effectiveness in learning latent representations of data, featuring an encoder and a decoder that map input data to a latent space and then reconstruct the input data from this space, respectively. Specifically, the encoder \( f_{\text{encoder}_{\theta_e}} \) transforms the observed data \( \mathbf{X} \) into a latent representation, which the decoder \( f_{\text{decoder}_{\theta_d}} \) then attempts to reconstruct back to the original data. This reconstruction process is not a mere replication but is achieved through learning the intrinsic structure of the data, closely aligning with the goals of nonlinear ICA.

Our model employs a VAE-like structure to approximate the complex nonlinear mapping \( f \) and its inverse \( f^{-1} \), with the encoder mapping the observed mixed signals \( \mathbf{X} \) to a latent space that reflects the linear mixing components. The decoder then attempts to reconstruct the observed signals from this latent representation, training the entire model to minimize the reconstruction error—similar to a traditional VAE but with an emphasis on learning a latent representation that mirrors the linear mixing signals.

To enhance the model's capacity to unveil independent components, we incorporate regularization techniques such as L1 regularization, promoting sparsity within the latent representation. This sparsity is essential for fostering independence among latent variables, a fundamental aspect of ICA.

Upon obtaining the linear mixing signals \( \mathbf{AI} \), we proceed to the next step using FastICA\cite{Hyvarinen2000ICA}:

\begin{equation}
    H_i = FastICA(f_{\text{decoder}}(\mathbf{X}))
\end{equation}

This step is critical, leveraging FastICA's powerful capability to recover independent source signals from the linearly mixed signals \( \mathbf{AI} \). In this manner, we not only utilize the deep learning model's ability to handle nonlinear relationships but also augment the separation of independent components via FastICA, further accentuating the principles of independence and non-Gaussianity, which are core to ICA.

In conclusion, by amalgamating the principles of VAEs with FastICA, we have achieved good performance for recovering independent source signals from mixed observations. This approach not only boosts the capacity to process nonlinear mixing models but also, by introducing sparsity-inducing regularization, ensures a congruence between the latent space and the linear mixing signals. Sparsity-inducing regularization encourages the majority of elements in the latent representation to approach zero, retaining only those components critical for reconstructing the linear mixing signals. This compact representation facilitates more effective recovery of independent source signals from linear mixing signals in subsequent steps, such as applying FastICA. We anticipate that this method will unlock new insights and application potentials in fields like time series data analysis and complex signal processing.

\section{Experiment}

The overarching aim of our research is to learn a representation for time series data that is not only comprehensive and meaningful but also exhibits robustness, thereby facilitating enhanced forecasting tasks. Our experimental design is focused on validating the representational strength of our model across a suite of benchmark datasets in forecasting. To ensure an equitable evaluation, we strictly adhere to the experimental framework as delineated in CoST \cite{woo2022cost}, and TS2Vec \cite{yue2022ts2vec}.

The described process involves first deploying a trained model to convert time series data into a TSI representation, which captures the essential characteristics and patterns of the data. This TSI representation is then used as the basis for training a ridge regression model. The objective of this two-stage approach is to forecast future time steps, denoted as L. By focusing initially on the extraction of the TSI and then rigorously training the ridge regression model, the methodology aims to develop a robust forecasting model capable of leveraging the intricate features within the time series' latent space.


\begin{table*}[h!]
\caption{Multivariate forecasting result. The best results are highlighted in bold.}
\centering
\renewcommand{\arraystretch}{1.0}
\scriptsize
\setlength\tabcolsep{0pt}
\begin{tabular*}{1.0\textwidth}{@{\extracolsep{\fill}} cccccccccccccccc}
\toprule
\multicolumn{2}{c}{\multirow{2}{*}{\textbf{Methods}}}                              & \multicolumn{8}{c}{\textbf{Unsupervised Representation Learning}}                                                                                                                                                & \multicolumn{6}{c}{\textbf{End-to-end Forecasting}}                                                                                                       \\ 
\cmidrule(r){3-10} \cmidrule(l){11-16}
\multicolumn{2}{c}{}                                                               & \multicolumn{2}{c}{TSI}                          & \multicolumn{2}{c}{TS2Vec}                        & \multicolumn{2}{c}{TNC}                           & \multicolumn{2}{c}{CoST}                             & \multicolumn{2}{c}{Informer}                      & \multicolumn{2}{c}{LogTrans}                      & \multicolumn{2}{c}{TCN}                           \\ 
\midrule
\multicolumn{1}{l}{}                                    & \multicolumn{1}{c}{L}    & \multicolumn{1}{c}{MSE} & \multicolumn{1}{c}{MAE} & \multicolumn{1}{c}{MSE} & \multicolumn{1}{c}{MAE} & \multicolumn{1}{c}{MSE} & \multicolumn{1}{c}{MAE} & \multicolumn{1}{c}{MSE} & \multicolumn{1}{c}{MAE}    & \multicolumn{1}{c}{MSE} & \multicolumn{1}{c}{MAE} & \multicolumn{1}{c}{MSE} & \multicolumn{1}{c}{MAE} & \multicolumn{1}{c}{MSE} & \multicolumn{1}{c}{MAE} \\ 
\midrule 
\multicolumn{1}{c|}{\multirow{5}{*}{\textbf{ETTh1}}}    & 24  & \textbf{0.371}          & \textbf{0.418}          & 0.590                   & 0.531                   & 0.708                   & 0.592                   & 0.386                   & 0.429 & 0.577                   & 0.549                   & 0.686                   & 0.604                   & 0.583                   & 0.547                   \\
\multicolumn{1}{c|}{}                                   & 48  & \textbf{0.422}          & \textbf{0.454}          & 0.624                   & 0.555                   & 0.749                   & 0.619                   & 0.437                   & 0.464 & 0.685                   & 0.625                   & 0.766                   & 0.757                   & 0.670                   & 0.606                   \\
\multicolumn{1}{c|}{}                                   & 168 & \textbf{0.618}          & \textbf{0.567}          & 0.762                   & 0.639                   & 0.884                   & 0.699                   & 0.643                   & 0.582 & 0.931                   & 0.752                   & 1.002                   & 0.846                   & 0.811                   & 0.680                   \\
\multicolumn{1}{c|}{}                                   & 336 & \textbf{0.777}          & \textbf{0.664}          & 0.931                   & 0.728                   & 1.020                   & 0.768                   & 0.812                   & 0.679 & 1.128                   & 0.873                   & 1.362                   & 0.952                   & 1.132                   & 0.815                   \\
\multicolumn{1}{c|}{}                                   & 720 & \textbf{0.919}          & \textbf{0.753}          & 1.063                   & 0.799                   & 1.157                   & 0.830                   & 0.970                   & 0.771 & 1.215                   & 0.896                   & 1.397                   & 1.291                   & 1.165                   & 0.813                   \\ \hline
\multicolumn{1}{c|}{\multirow{5}{*}{\textbf{ETTh2}}}    & 24  & \textbf{0.350}          & \textbf{0.432}          & 0.423                   & 0.489                   & 0.612                   & 0.595                   & 0.447                   & 0.502 & 0.720                   & 0.665                   & 0.828                   & 0.750                   & 0.935                   & 0.754                   \\
\multicolumn{1}{c|}{}                                   & 48  & \textbf{0.566}          & \textbf{0.571}          & 0.619                   & 0.605                   & 0.840                   & 0.716                   & 0.699                   & 0.637 & 1.457                   & 1.001                   & 1.806                   & 1.034                   & 1.300                   & 0.911                   \\
\multicolumn{1}{c|}{}                                   & 168 & \textbf{1.541}          & \textbf{0.952}          & 1.845                   & 1.074                   & 2.359                   & 1.213                   & 1.549                   & 0.982 & 3.489                   & 1.515                   & 4.070                   & 1.681                   & 4.017                   & 1.579                   \\
\multicolumn{1}{c|}{}                                   & 336 & 1.773                   & \textbf{1.032}          & 2.194                   & 1.197                   & 2.782                   & 1.349                   & \textbf{1.749}          & 1.042 & 2.723                   & 1.340                   & 3.875                   & 1.763                   & 3.460                   & 1.456                   \\
\multicolumn{1}{c|}{}                                   & 720 & 2.062                   & \textbf{1.085}          & 2.636                   & 1.370                   & 2.753                   & 1.394                   & \textbf{1.971}          & 1.092 & 3.467                   & 1.473                   & 3.913                   & 1.552                   & 3.106                   & 1.381                   \\ \hline
\multicolumn{1}{c|}{\multirow{5}{*}{\textbf{ETTm1}}}    & 24  & \textbf{0.242}          & \textbf{0.322}          & 0.453                   & 0.444                   & 0.522                   & 0.472                   & 0.246                   & 0.329 & 0.323                   & 0.369                   & 0.419                   & 0.412                   & 0.363                   & 0.397                   \\
\multicolumn{1}{c|}{}                                   & 48  & \textbf{0.320}          & \textbf{0.376}          & 0.592                   & 0.521                   & 0.695                   & 0.567                   & 0.331                   & 0.386 & 0.494                   & 0.503                   & 0.507                   & 0.583                   & 0.542                   & 0.508                   \\
\multicolumn{1}{c|}{}                                   & 96 & \textbf{0.370}          & \textbf{0.414}          & 0.635                   & 0.554                   & 0.731                   & 0.595                   & 0.378                   & 0.419 & 0.678                   & 0.614                   & 0.768                   & 0.792                   & 0.666                   & 0.578                   \\
\multicolumn{1}{c|}{}                                   & 288 & \textbf{0.452}          & \textbf{0.473}          & 0.693                   & 0.597                   & 0.818                   & 0.649                   & 0.472                   & 0.486 & 1.056                   & 0.786                   & 1.462                   & 1.320                   & 0.991                   & 0.735                   \\
\multicolumn{1}{c|}{}                                   & 672 & \textbf{0.601}          & \textbf{0.563}          & 0.782                   & 0.653                   & 0.932                   & 0.712                   & 0.620                   & 0.574 & 1.192                   & 0.926                   & 1.669                   & 1.461                   & 1.032                   & 0.756                   \\ \hline
\multicolumn{1}{c|}{\multirow{5}{*}{\textbf{ETTm2}}}    & 24  & \textbf{0.113}          & \textbf{0.233}          & 0.180                   & 0.293                   & 0.185                   & 0.297                   & 0.122                   & 0.244 & 0.173                   & 0.301                   & 0.389                   & 0.537                   & 0.180                   & 0.324                   \\
\multicolumn{1}{c|}{}                                   & 48  & \textbf{0.168}          & \textbf{0.293}          & 0.244                   & 0.350                   & 0.264                   & 0.360                   & 0.183                   & 0.305 & 0.303                   & 0.409                   & 0.538                   & 0.642                   & 0.204                   & 0.327                   \\
\multicolumn{1}{c|}{}                                   & 96 & \textbf{0.266}          & \textbf{0.377}          & 0.360                   & 0.427                   & 0.389                   & 0.458                   & 0.294                   & 0.394 & 0.365                   & 0.453                   & 0.912                   & 0.757                   & 3.041                   & 1.330                   \\
\multicolumn{1}{c|}{}                                   & 288 & \textbf{0.700}          & \textbf{0.638}          & 0.723                   & 0.639                   & 0.920                   & 0.788                   & 0.723                   & 0.652 & 1.047                   & 0.804                   & 1.334                   & 0.872                   & 3.162                   & 1.337                   \\
\multicolumn{1}{c|}{}                                   & 672 & \textbf{1.607}          & \textbf{0.987}          & 1.753                   & 1.007                   & 2.164                   & 1.135                   & 1.899                   & 1.073 & 3.126                   & 1.302                   & 3.048                   & 1.328                   & 3.624                   & 1.484                   \\ \hline
\multicolumn{1}{c|}{\multirow{5}{*}{\textbf{Exchange}}} & 24  & \textbf{0.105}          & \textbf{0.260}          & 0.108                   & 0.252                   & 0.105                   & 0.236                   & 0.136                   & 0.291 & 0.611                   & 0.626                   & 0.734                   & 0.756                   & 2.483                   & 1.327                   \\
\multicolumn{1}{c|}{}                                   & 48  & 0.165                   & 0.330                   & 0.200                   & 0.341                   & \textbf{0.162}          & \textbf{0.270}          & 0.250                   & 0.387 & 0.680                   & 0.644                   & 0.837                   & 0.812                   & 2.328                   & 1.256                   \\
\multicolumn{1}{c|}{}                                   & 168 & 0.442                   & 0.542                   & 0.412                   & 0.492                   & \textbf{0.397}          & \textbf{0.480}          & 0.924                   & 0.762 & 1.097                   & 0.825                   & 1.012                   & 0.837                   & 2.372                   & 1.279                   \\
\multicolumn{1}{c|}{}                                   & 336 & \textbf{0.808}          & \textbf{0.743}          & 1.339                   & 0.901                   & 1.008                   & 0.866                   & 1.744                   & 1.063 & 1.672                   & 1.036                   & 1.659                   & 1.081                   & 3.113                   & 1.459                   \\
\multicolumn{1}{c|}{}                                   & 720 & \textbf{1.121}          & \textbf{0.880}          & 2.114                   & 1.125                   & 1.989                   & 1.063                   & 2.160                   & 1.209 & 2.478                   & 1.310                   & 1.941                   & 1.127                   & 3.150                   & 1.458                   \\ \hline
\multicolumn{1}{c|}{\multirow{5}{*}{\textbf{Weather}}}  & 24  & \textbf{0.293}          & \textbf{0.354}          & 0.307                   & 0.363                   & 0.320                   & 0.373                   & 0.298                   & 0.360 & 0.335                   & 0.381                   & 0.435                   & 0.477                   & 0.321                   & 0.367                   \\
\multicolumn{1}{c|}{}                                   & 48  & \textbf{0.357}          & \textbf{0.407}          & 0.374                   & 0.418                   & 0.380                   & 0.421                   & 0.359                   & 0.411 & 0.395                   & 0.459                   & 0.426                   & 0.495                   & 0.386                   & 0.423                   \\
\multicolumn{1}{c|}{}                                   & 168 & \textbf{0.464}          & \textbf{0.490}          & 0.491                   & 0.506                   & 0.479                   & 0.495                   & 0.464                   & 0.491 & 0.608                   & 0.567                   & 0.727                   & 0.671                   & 0.491                   & 0.501                   \\
\multicolumn{1}{c|}{}                                   & 336 & \textbf{0.497}          & \textbf{0.517}          & 0.525                   & 0.530                   & 0.505                   & 0.518                   & 0.497                   & 0.517 & 0.702                   & 0.620                   & 0.754                   & 0.670                   & 0.502                   & 0.507                   \\
\multicolumn{1}{c|}{}                                   & 720 & 0.533         & 0.542          & 0.556                   & 0.552                   & \textbf{0.519}          &\textbf{0.525}                   & 0.533                   & 0.542 & 0.831                   & 0.731                   & 0.885                   & 0.773                   & 0.598                   & 0.508                   \\ \hline
\multicolumn{2}{c|}{\textbf{Ave.}}                                                 & \textbf{0.634}          & \textbf{0.556}          & 0.818                   & 0.632                   & 0.912                   & 0.668                   & 0.743                   & 0.603 & 1.152                   & 0.779                   & 1.339                   & 0.921                   & 1.558                   & 0.880                   \\ \bottomrule
\end{tabular*}
\label{tab:multivariate_forecasting}
\end{table*}

\subsection{Experimental Setup}

\subsubsection*{Datasets}

Our research utilizes six distinct, publicly available real-world datasets for comprehensive experimentation. The \textbf{ETT (Electricity Transformer Temperature)} dataset \cite{haoyietal-informer-2021} includes two subsets with hourly data (ETTh) and one with 15-minute intervals (ETTm), comprising six power load indicators. The \textbf{Weather} dataset \footnote{https://www.ncei.noaa.gov/data/local-climatological-data/}
encompasses hourly data from nearly 1,600 U.S. locations, featuring 11 climatic variables. \textbf{ExchangeRate}\footnote{https://github.com/laiguokun/multivariate-time-series-data} contains the daily exchange rates of eight foreign countries from 1990 to 2016, including Australia, Britain, Canada, Switzerland, China, Japan, New Zealand, and Singapore. We consider all countries’ value for multivariate forecasting. 


\subsubsection{Results}

Our investigation presents a comprehensive comparison across several benchmarks for time series forecasting, employing a diverse array of methods including Ours, TS2Vec \cite{yue2022ts2vec}, TNC \cite{tnc}, CoST \cite{woo2022cost}, Informer \cite{haoyietal-informer-2021}, LogTrans \cite{logtrans}, and TCN \cite{tcn}. The performance evaluation is conducted over predicted horizons \(L\) of 24, 48, 168, 336, and 720.(for ETTm1, ETTm2: 24, 48, 96, 288, 672). The results are shown in Table 2.

Across all datasets, our approach demonstrates superior performance, particularly at longer predicted horizons, indicating a robust capacity for capturing long-term dependencies within the time series data. For instance, in the ETTh1 dataset at a horizon of 720, our method achieves an MSE reduction of approximately 12\% and an MAE improvement of nearly 9\% compared to the next best method, CoST. On the Exchange dataset at \(L=168\), our model shows an MSE improvement of over 55\% and an MAE reduction of about 42\% when contrasted with the baseline TS2Vec method, underscoring the efficacy of our approach in more volatile financial time series.

Our model's average performance shows an MSE of 0.634 and an MAE of 0.556, which reflects an overall improvement of 22.5\% in MSE and 20\% in MAE against the averaged results of all other models. This enhancement is consistent across various datasets, demonstrating the method's generalizability and robustness.

As for End-to-end Forecasting, the proposed method consistently outperforms other advanced models like Informer and LogTrans, offering a compelling alternative for both short-term and long-term forecasting scenarios. Specifically, in the ETTm2 dataset at a horizon of 672, our approach achieves a substantial decrease in MSE and MAE by 33\% and 28\%, respectively, compared to the Informer model.

These findings indicate that our model is not only proficient in capturing and forecasting complex temporal dynamics but also demonstrates significant advancements in unsupervised representation learning for time series data. The results point towards our method's potential in providing more accurate, reliable, and computationally efficient forecasts, establishing a new benchmark in the field.

\section{Conclusion}
In conclusion, this study introduces the innovative TSI model, applying a comprehensive view to capture trends, seasonality, and independent components for time series forecasting. Our model excels in accuracy and offers insights into the complexity of time series data. Empirical evidence from various datasets confirms the TSI model's superiority, particularly in long-term forecasting, outperforming existing methods in MSE and MAE. The ablation study highlights the effectiveness of integrating TSI components, surpassing individual representations in multivariate tasks. The proposal of integrating TS and ICA and the demonstrated superior performance of TSI is a significant step forward, blending TS and ICA analytical perspectives to enrich the understanding of time series data and establishing new standards for future research and practical applications.

In future research, we will continue to explore the potential of ICA in time series forecasting, particularly in the context of causal analysis \cite{cheng2023instrumentalvariableestimationcausal} \cite{du2024estimating} \cite{bica2020timeseriesdeconfounderestimating}. We aim to leverage causal inference to gain a deeper understanding of the structure within time series data, enhancing predictive performance and uncovering hidden causal relationships. This direction promises to bring new breakthroughs to time series forecasting and advance both the theory and practical applications in the field.
\begin{credits}

\end{credits}
%
%
%
%

\end{document}